\documentclass[sigconf]{acmart}

\usepackage{graphicx}
\usepackage{amsmath}
\usepackage{array}
\usepackage{booktabs}
\usepackage{multirow}
\usepackage{arydshln}
\usepackage{upgreek}
\usepackage[shortlabels]{enumitem}

\newcolumntype{C}[1]{>{\centering\arraybackslash}m{#1}}

\AtBeginDocument{%
  \providecommand\BibTeX{{%
    \normalfont B\kern-0.5em{\scshape i\kern-0.25em b}\kern-0.8em\TeX}}}

\copyrightyear{2021}
\acmYear{2021}
\setcopyright{acmcopyright}
\acmConference[CIKM '21]{Proceedings of the 30th ACM International Conference on Information and Knowledge Management}{November 1--5, 2021}{Virtual Event, QLD, Australia}
\acmBooktitle{Proceedings of the 30th ACM International Conference on Information and Knowledge Management (CIKM '21), November 1--5, 2021, Virtual Event, QLD, Australia}
\acmPrice{15.00}
\acmDOI{10.1145/3459637.3482346}
\acmISBN{978-1-4503-8446-9/21/11}

\begin{document}
\fancyhead{}

\title{Generating Compositional Color Representations from Text}

\author{Paridhi Maheshwari}
\authornote{Work done when authors were at Adobe Research}
\affiliation{\institution{Stanford University}}
\email{paridhi@stanford.edu}

\author{Nihal Jain}
\authornotemark[1]
\affiliation{\institution{Carnegie Mellon University}}
\email{nihalj@cs.cmu.edu}

\author{Praneetha Vaddamanu}
\affiliation{\institution{Adobe Research}}
\email{vaddaman@adobe.com}

\author{Dhananjay Raut}
\authornotemark[1]
\affiliation{\institution{Adobe}}
\email{raut@adobe.com}

\author{Shraiysh Vaishay}
\authornotemark[1]
\affiliation{\institution{Advanced Micro Devices Inc.}}
\email{shraiysh.vaishay@amd.com}

\author{Vishwa Vinay}
\affiliation{\institution{Adobe Research}}
\email{vinay@adobe.com}

\renewcommand{\shortauthors}{Maheshwari et al.}

\begin{abstract}
We consider the cross-modal task of producing color representations for text phrases. Motivated by the fact that a significant fraction of user queries on an image search engine follow an (attribute, object) structure, we propose a generative adversarial network that generates color profiles for such bigrams. We design our pipeline to learn composition - the ability to combine seen attributes and objects to unseen pairs. We propose a novel dataset curation pipeline from existing public sources. We describe how a set of phrases of interest can be compiled using a graph propagation technique, and then mapped to images. While this dataset is specialized for our investigations on color, the method can be extended to other visual dimensions where composition is of interest. We provide detailed ablation studies that test the behavior of our GAN architecture with loss functions from the contrastive learning literature. We show that the generative model achieves lower Fr\'echet Inception Distance than discriminative ones, and therefore predicts color profiles that better match those from real images. Finally, we demonstrate improved performance in image retrieval and classification, indicating the crucial role that color plays in these downstream tasks.
\end{abstract}



\maketitle

\section{Introduction}
We consider the problem of cross-modal retrieval where textual queries are used to produce a ranked list of images. User queries can be very diverse and exhibit a range of linguistic structures. In addition, the richness and ambiguities of language make the accurate retrieval of images a challenging task. The ranker needs to incorporate multiple relevance indicators for queries as well as items. In this paper, we place constraints on both these aspects - we focus on (a) queries with an attribute-object bigram structure, and (b) the role of color as a relevance ranking feature.

Attribute-Object pairs are a commonly observed structure in search queries, e.g., `cute dog' or `happy child'. Since we consider the domain of image retrieval, we are interested in how these phrases manifest visually. The characteristics of the combined phrase are a composition of the visual intent of the attribute (or adjective) and the object (or noun). Modeling the constituent terms separately, and knowing how to combine them, helps us generalize to new concepts~\cite{misra2017red}, e.g., `happy dog'. Such an improved understanding of the queries enables the building of a more robust search ranker. In addition, we might be able to support higher-level intents formed by the same building blocks, such as complex composite queries like `red bricks on a white background’.

Specifically, we focus on the impact of attribute-object compositionality on color. Our focus is motivated by the central role that color plays in image processing and retrieval~\cite{plataniotis2013color, hsu1995integrated, jain1996image}. In addition, there is an inherent color intent associated with several user queries~\cite{maheshwari2020learning}. For example, while queries such as `raw apple’ and `deep sea’ do not have explicit color mentions, they evoke specific color semantics that may be useful to infer image relevance. We take steps towards this by modeling attribute-object compositionality in user queries to generate color representations.

Modeling compositionality requires that the dataset cover the set of combinations. That is, for $\mathcal{N}_a$ attributes and $\mathcal{N}_o$ objects, the set of (attribute, object) bigrams in the dataset should ideally include the all $\mathcal{N}_a \times \mathcal{N}_o$ combinations with sufficient image examples. The individual terms might have varying degrees of color intent (e.g. `dark' versus `happy' as attributes). The corresponding bigrams might therefore also vary in how they impact color (e.g. `dark sea` $~>~$ `dark sky` $~>~$ `happy sky`), but might also lead to unlikely pairs (e.g. `happy sea'). To handle these scenarios, we design a novel pipeline to curate a dataset of images with their corresponding (attribute, object) pair labels. Our approach enables us to capture a diverse set of phrases with high color intent. Our dataset creation strategy could be potentially useful for other studies into compositionality along specific axes (in our case, color). Our dataset construction includes a visual attention mechanism to ground the (attribute, object) pairs into images. This enables us to extract cleaner and more relevant color profiles from images, conditioned on the bigram.

We propose an adversarial learning approach~\cite{goodfellow2014generative} for generating color representations compositionally from textual input. Inspired by its recent success in several domains, we experiment with different loss functions from the contrastive learning paradigm for training our GAN. The generator takes word embeddings of the attribute and object as input, this aids the model in generalizing in the text modality. Since the perception of color is naturally rooted in the visual domain, we wish to provide the model with information from images even though our end-task only utilizes textual input for generating color profiles. To enable this, our model takes the image modality as input while training the discriminator. The coupled training between the generator and the discriminator, therefore, leads to the visual modality affecting the final model, while the generator utilizes only text at test time.

Our evaluation strategy is trifold: (i) assess the quality of the generated color representations for input (attribute, object) pairs; (ii) the quality of our conditional GAN architecture and training pipeline; and (iii) the effectiveness of introducing color features in downstream tasks. First, we compare the performance of our text-to-color encoder against a discriminative baseline that predicts color profiles from (attribute, object) word embeddings. We evaluate the models using $L2$ distances between the generated and ground-truth color profiles and show that our generative model outperforms the baseline. Second, we compute the Fr\'echet Inception Distance~\cite{fid} between the predicted and real color profiles to evaluate the quality of our generative model and discuss a series of ablation experiments to study the effectiveness of components of the GAN objective. Finally, we demonstrate the usefulness of color as a feature for cross-modal retrieval and image classification.

We summarize our key contributions as follows:
\begin{enumerate}[topsep=0pt]
    \item We present a strategy to prepare text-to-color datasets from existing public sources. The text phrases are limited to (attribute, object) bigrams to study color compositionality.
    \item We propose a generative adversarial modeling approach to produce color representations of textual phrases. This model takes visual cues from the image modality at train time but does not require these image features at test time.
    \item We perform a comparative study of loss functions adapted from contrastive learning literature for the task of text-to-color encoding using GANs.
    \item Finally, by using the generated color representations of textual queries as features for image ranking, we demonstrate that search relevance can be improved.
\end{enumerate}

\section{Related Work}
In this section, we review prior work related to the compositional structure of language and obtaining color representations from text.

\subsection{Language and Color}
The richness of language in being able to describe complex visual features has been a long-studied subject~\cite{regier2009language}. Various psychological studies~\cite{winawer2007russian, roberson2000color} have also demonstrated the strong association between colors and natural language phrases. Certain words like \textcolor{cyan}{water} and \textcolor{red}{rose} exemplify this association. 

To study this relation between color and language, several datasets have been curated that provide a mapping between the modalities. The XKCD dataset~\cite{munroe_2010} labels textual phrases with colors, and was setup via a crowd-sourced survey. \cite{white2017learning} use this dataset to learn a probability distribution in HSV color space conditioned on the name of the color. \cite{setlur2015linguistic} also employ the XKCD dataset along with Google's n-gram corpus~\cite{michel2011quantitative} to prepare phrases with high color association. Google's n-gram corpus was also used by~\cite{lindner2013automatic} to select commonly used single words for use in the task of color palette extraction. We also rely on Google's n-gram dataset for finding color related words, as described in section~\ref{Dataset}, but we describe a new mechanism to generate textual queries that not only have high color intent but also have an (attribute, object) structure.

Several research efforts have attempted to arrive at mappings between the text and color. \cite{kawakami2016character} use a character-level LSTM to predict a color given a name. Other works~\cite{bahng2018coloring, lindner2013automatic} focus on arriving at color palettes from textual input. In contrast, we focus on the task of generating color histograms that convey much richer information and have greater utility in a cross-modal search as an additional feature embedding. While ~\cite{maheshwari2020learning} also focus on generating color histograms from text, our work focuses specifically on the compositional structure in language -- which we argue is crucial for generating relevant color representations. We propose a generative model for text to color, also utilizing the image modality which contains crucial information for color intent.

\subsection{Composition and Context}
According to the compositionality principle observed in language, novel concepts can be constructed from primitive building blocks. Following \cite{nagarajan2018attributes, Li_2020_CVPR, purushwalkam2019task}, we model compositionality by treating attributes and objects as primitives. This intuitive principle is closely connected with the principle of contextuality which states that the behaviour of a primitive varies in the presence of others~\cite{misra2017red}. Specifically, the same attribute can affect different objects in different ways and the same object elicits different behaviours when modified by different attributes. For example, \textit{'ripe'} when used in context of 'apple' has a different visual manifestation than when \textit{'ripe'} is used in context of \textit{'mango'}. Similarly, the object \textit{'car'} evokes different intuitions when it is modified by \textit{'sporty'} and \textit{'old'}. This interaction between compositionality and contextuality has recently been a subject of study in various fields. Specifically, in recent machine learning literature~\cite{purushwalkam2019task, nagarajan2018attributes, tokmakov2019learning, nan2019recognizing}, these concepts have formed the basis for zero-shot learning or few-shot learning where generalization is achieved by modeling compositions of primitives which are not part of the training set.

In the present work, we explore composing attributes and objects in textual queries to extract color representations that can be used for retrieving relevant images. We further demonstrate the use of our method to compose unseen combinations of attributes and objects to derive intuitive color representations.

\subsection{Contrastive Learning}
Contrastive learning approaches to representation learning have recently gained traction due to their success in several domains such as computer vision and natural language processing~\cite{chen2020simple,fang2020cert,le2020contrastive, maheshwari2021scene}. The intuition behind these approaches is to bring similar pairs of data points (typically referred to as the anchor and the positive) closer to each other than dissimilar pairs (anchor and negative) in an embedding space. This is achieved through non-linear transformations learnt using objectives like the triplet loss~\cite{weinberger2006distance,schroff2015facenet} or the InfoNCE objective~\cite{oord2018representation} amongst others. The authors of ~\cite{wei2019adversarial,nan2019recognizing,tokmakov2019learning} make use of the contrastive learning framework to model attribute-object tasks by exploiting the compositional nature of the inputs to sample positives and negatives for an anchor data point. These loss functions above operate on a triple of data points - the anchor, positive, and the negative. ~\cite{wei2019adversarial} defines a quintuplet loss, an extension of the triplet loss that introduces intuitions from compositionality into the contrastive learning setting. 

Building on these approaches, we adapt the contrastive learning framework to generate color representations that respect compositionality and context. This is achieved by using the notions of similar and dissimilar pairs as described in these works to drive our sampling strategies while training our machine learning models. As an indicator of the intuition behind these methods, for a given anchor example image for the (attribute, object) bigram $(A,O)$, positive examples are all other images tagged with the same bigram. Negative examples can also be simply obtained as all images not associated with this bigram. However, bigrams that share either the attribute or the object (i.e., $(A,X)$ or $(Y,O)$) are partially related. These relative preferences amongst images associated with these classes are naturally exploited by the contrastive loss formulations. 

\section{Dataset} \label{Dataset}
We leverage image datasets labeled with (attribute, object) pair information to develop algorithms that generate color profiles from text. It is of primary importance to ensure a rich and diverse set of (attribute, object) phrases which are not limited to trivial color mentions (such as \textit{`red scarf'} or \textit{`blue sky'}), but also include implicit and inherent indicators (such as \textit{`cranberry juice'} or \textit{`deep sea'}). While there exist public datasets on object transformations~\cite{isola2015discovering, yu2014fine}, they attend to variations in physical state and appearance (for example, \textit{rope} can be \textit{thin}, \textit{short}, \textit{coiled}). In the current work, we focus on one particular visual aspect, i.e., color, and propose a novel approach to curate datasets that specifically capture (attribute, object) phrases with high color intent.

\subsection{Curation}
We start by gathering the set of commonly occurring (attribute, object) phrases from textual n-grams. The bigram corpus from Google’s n-gram dataset~\cite{michel2011quantitative} contains the list of all contiguous sequences of $2$ words present in the Google corpus along with their frequency count. Based on the linguistic type of the constituent words, we extract all phrases where the first word is an adjective (attribute) and second is a noun (object). To remove non-visual concepts (such as \textit{`old wisdom'} or \textit{`European community'}), we restrict our vocabulary using well known lists of concrete nouns~\cite{brysbaert2014concreteness} and descriptive adjectives~\cite{descriptive2020adjectives}. This approach results in the set of frequently occurring visual concepts in public corpora.

Given our specific focus on color, we would like to exclude phrases assumed to have no color intent (such as \textit{`epithelial cells'} or \textit{`electric fields'}). To achieve this, we build a bipartite graph between attributes and objects and utilize a hopping logic to iteratively select pairs. Starting with the $11$ Basic Color Terms~\cite{berlin1991basic} as attributes, we obtain the list of objects that occur most frequently (top $f$) with this set of seed colors. In the next step, we identify the attributes that they most commonly occur alongside. This completes one traversal, termed as a single hop of the bipartite graph. The selection process repeats with multiple hops $h$ till the required number of (attribute, object) pairs have been selected.

For the model to learn compositionality of attributes and objects, we need to ensure sufficient occurrences of every word and we achieve this by maintaining a threshold $t_a$ and $t_o$ for the number of unique attributes per object and unique objects per attribute respectively. Lastly, we fetch images for every (attribute, object) pair by querying the Google Image Search engine and retrieving the top results. The statistics of the final dataset are summarized in Table~\ref{datasetStatistics}.

\begin{table}[!h]
\caption{Statistics of the (attribute, object, image) datasets created using the Google Bigrams corpus.}
\label{datasetStatistics}
\centering
\begin{tabular}{r C{1.5cm} }
    \toprule
    \# attributes & $130$ \\
    \# objects & $211$ \\
    \# pairs & $1460$ \\
    \# images per pair & $33.55$ \\
    \# images & $48983$ \\
    \bottomrule
\end{tabular}
\end{table}

We essentially use the distance from Basic Color Terms~\cite{berlin1991basic} in the bipartite graph as a proxy for color intent. For example, starting with color terms `red' and `blue' as attributes gives us the frequent pairs `red rose', `blue sea' and so on. Now, using `rose' and `sea' as seed input, we fetch the pairs `wild rose', `deep sea' and `stormy sea'. As the number of hops increases, the phrases become more generic and less color-centric. This technique can be generalized to other visual properties such as texture, emotions, aesthetics -- by choosing an appropriate seed set of adjectives/attributes.

\subsection{Color Representation for Images}
Computational pipelines for color leverage well-known models and representations~\cite{wyszecki1982color}. In this work, the downstream application (image retrieval and classification) dictates the choice of color space and distance functions. Since our task involves human perception and interpretation, we utilize the LAB space which is known to be perceptually uniform -- distances in LAB space correspond to similar visually perceived changes in color. 

We divide the range spanned by the $3$ axes uniformly to create discrete bins. And a given pixel is mapped to one of the bins. And finally, the image is represented as a histogram over the bins such that each bar in the histogram is proportional to the fraction of pixels belonging to that bin. Note that utilizing larger bin widths leads to image level histograms that are less sparse, but with the disadvantage of having lost the detail. The choice of bin sizes, therefore, needs to trade-off the informativeness of fine-grained representations with the more robust coarse discretizations.  

The discretization itself introduces some noise into the representation, this can be partially alleviated by considering multiple bin widths. Specifically, we utilize two choices for the number of bins along (L, A, B) axes respectively -- $(9, 7, 8)$ and $(10, 10, 10)$. This gives us two separate histograms, each of sizes $9*7*8 = 504$ and $10*10*10 = 1000$ elements. Concatenating these leads to a combined $1504$-dimensional color embedding for an image. The specific choice of bin widths and number of alternative discretizations are design choices. In the current paper, we show results for a standard configuration, focussing on the central problem of interest -- the building of a generative text-to-color model.

\begin{figure}[!h]
\centering
\includegraphics[width=\columnwidth]{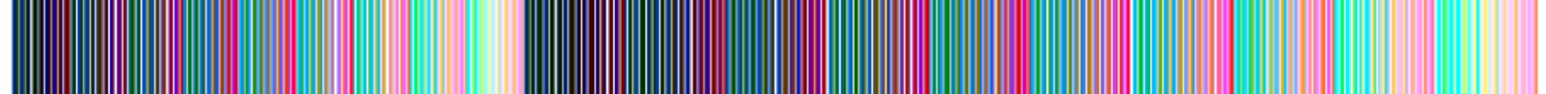}
\caption{Final set of 1504 color bins obtained by uniformly quantizing the LAB space. Note the repeating trend of the first 504 and the last 1000 bins, a result of concatenating histograms from two different LAB space divisions.}
\label{colourBins}
\end{figure}

\begin{figure}[!h]
\centering
\includegraphics[width=\columnwidth]{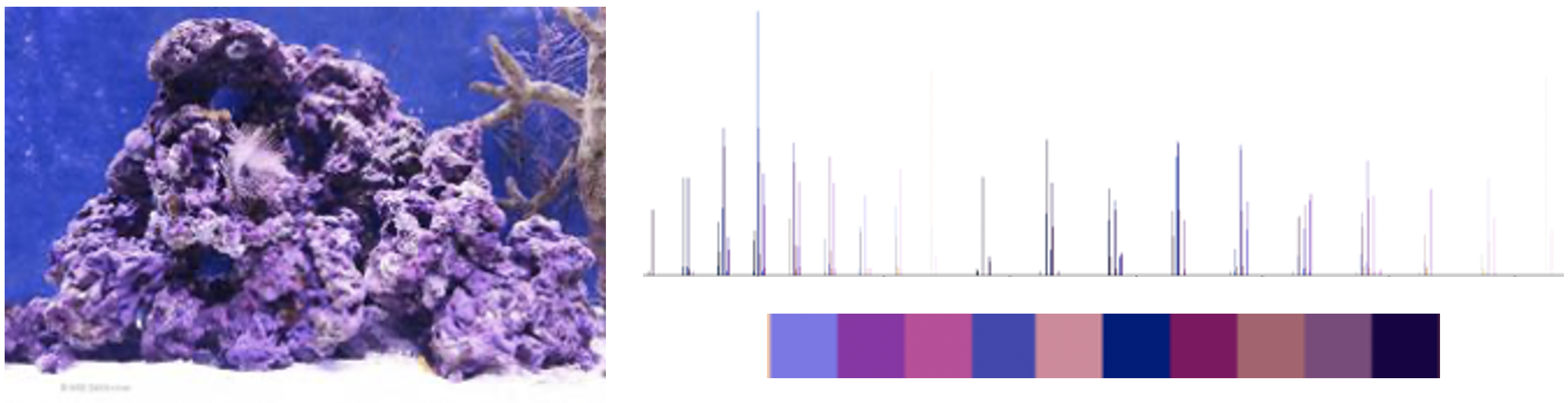}
\caption{Sample image, its color histogram and palette. It is evident that the purple and blue bins have the highest peaks while colors like brown have fired in smaller contributions.}
\label{imageAndColourRep}
\end{figure}

Figure~\ref{colourBins} provides a visualization of the $1504$ color bins and Figure~\ref{imageAndColourRep} shows an image of \textit{`coralline sea'} and its color histogram -- the height of the bar represents the weight of the corresponding bin, and the color corresponds to its LAB value. For easier interpretation, we extract a representative palette from the histograms by clustering similar shades together and sampling from the result. This generates a diverse summary that captures the majority shades from the original histogram. We will use the palette as an intuitive visualization for the output of our models.

\subsection{Modeling Visual Attention}

\begin{figure}[!h]
\centering
\includegraphics[width=0.8\columnwidth]{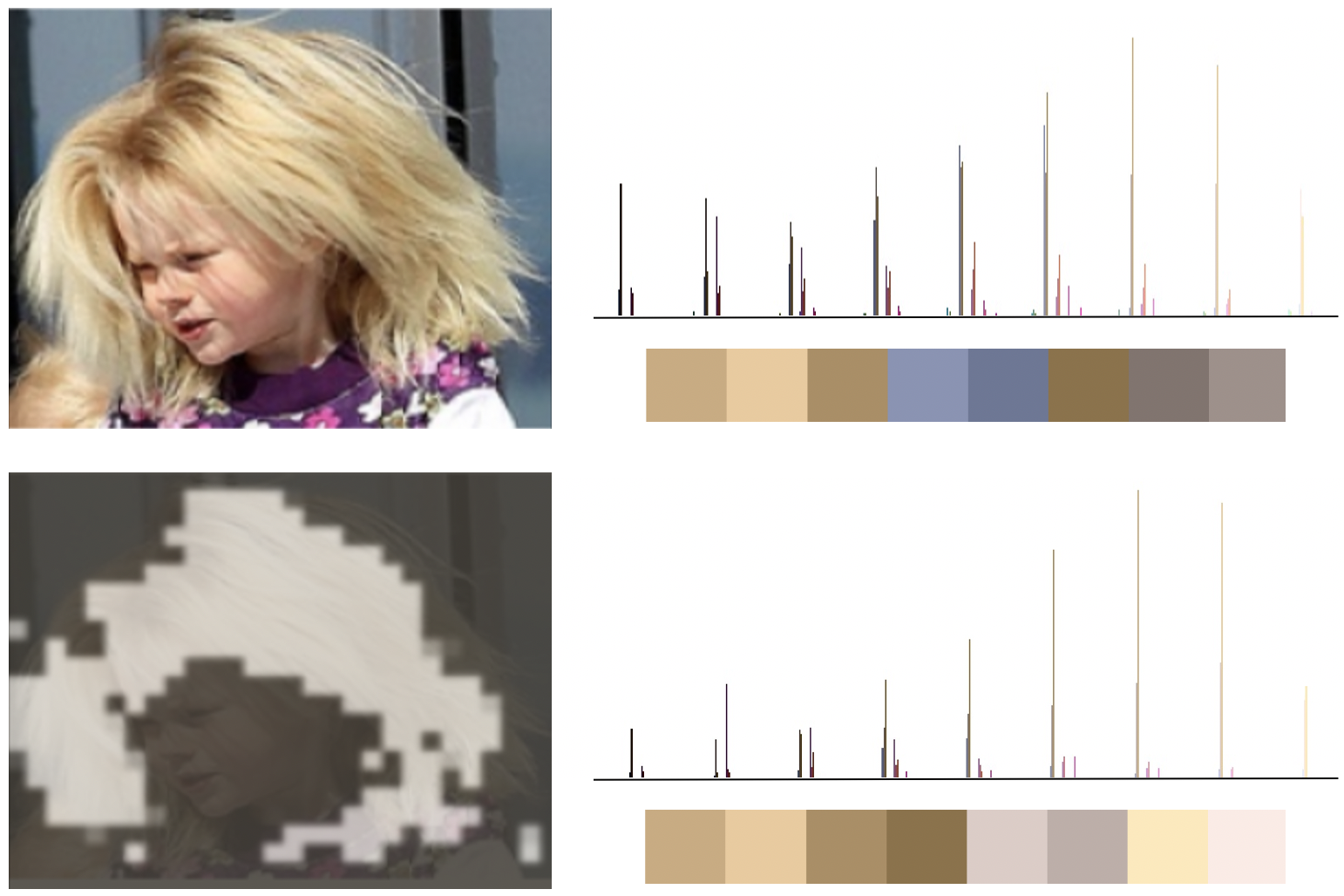}
\caption{For the pair \textit{`blond hair'}, we use attention map to identify relevant parts of the image and produce a less-noisy color representation with peaks towards blond and ignoring the blue from the irrelevant parts of the image.}
\label{imageAndAttention}
\end{figure}

We obtain images by querying Google's image search engine with the final set of unique (attribute, object) pairs, and utilize the technique described in the previous section to get color histograms for all images to be used for training our text-to-color models. This color representation gives uniform importance to all pixels in the image. However, conditioned on the text phrase, parts of the image may be more relevant than others, and we would like this intuition to affect the color representation appropriately. In order to identify relevant parts and extract cleaner color representations, we train a Convolutional Neural Network~\cite{jetley2018learn} on the classification task which internally uses visual attention to focus on parts of images. The model takes an image as input and predicts the attribute and object, while simultaneously learning an attention map over the image. We use the normalized attention weights from the trained model to give differential importance to individual pixels and create better color profiles. This is illustrated in Figure~\ref{imageAndAttention}.

\begin{figure}[!h]
\centering
\includegraphics[width=\columnwidth]{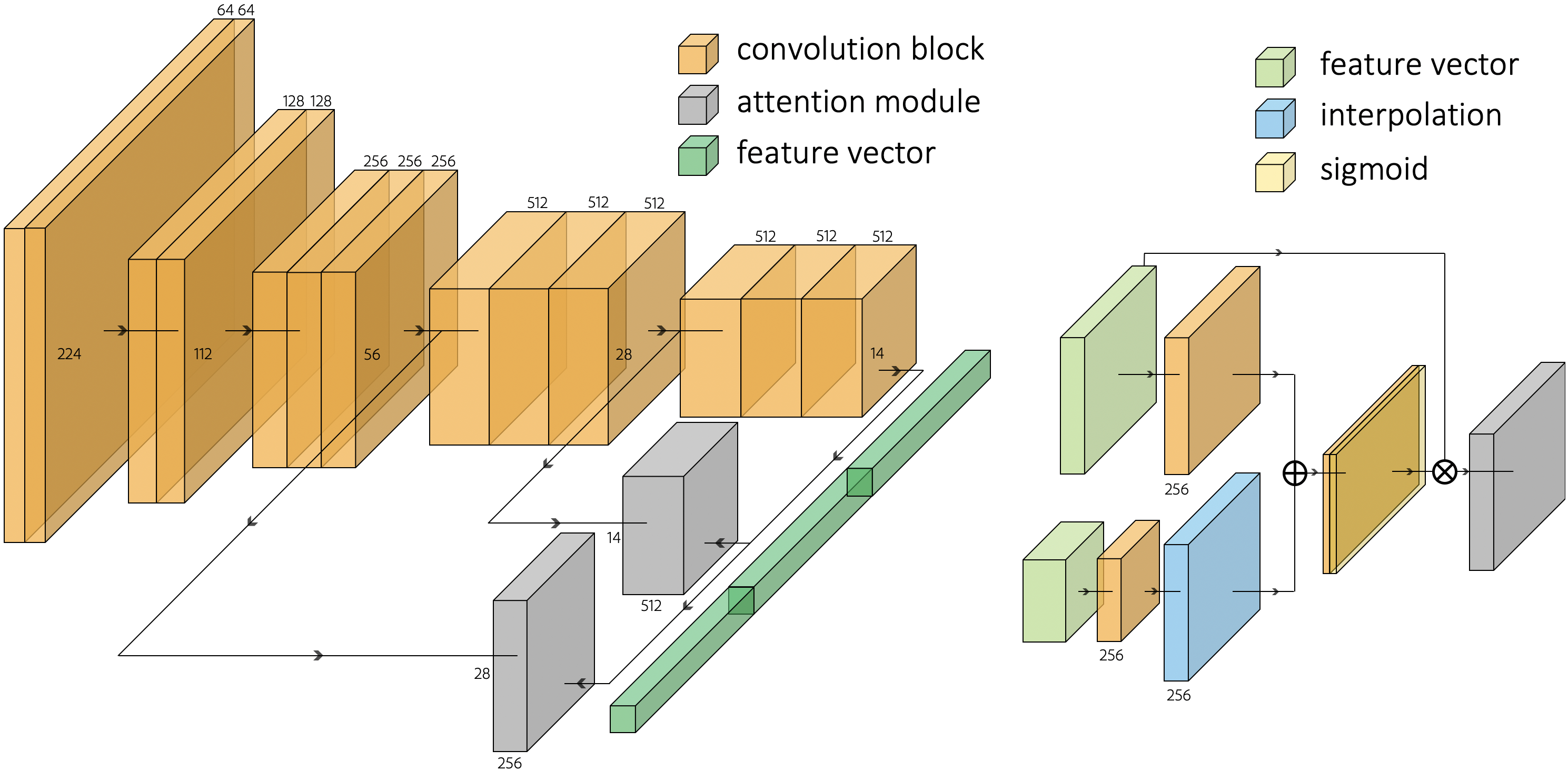}
\caption{Overall network architecture for learning visual attention (left) and individual attention module (right).}
\label{attentionModel}
\end{figure}

The model architecture is summarized in Figure~\ref{attentionModel} (left). The backbone is a VGG-16 network~\cite{simonyan2014very} with max pooling, ReLU activation and no dense layers. Two attention modules are applied at different intermediate stages which learn a pixel-wise attention map over the image. The learnt attention weights and global features are finally average pooled to get the feature vectors. The concatenated features are passed through two different classifiers, one for predicting attribute and the other for object respectively. Each classifier is a fully connected layer that computes confidence scores for all candidate classes (set of all objects or attributes). The model is trained using cross-entropy loss on one-hot encoded labels for both attribute and object given an input image. Lastly, we extract the spatial attention map from this model and perform a pixel-wise multiplication to obtain weighted color representations.

The individual attention modules, shown in Figure~\ref{attentionModel} (right), are a function of both intermediate representations as well as global image features. After passing through separate convolution layers, the global features are upsampled using bilinear interpolation to align spatial size to that of the input image. This is followed by an element-wise addition with intermediate features to get an attention map. The output of the attention module is an attention weighted feature space, i.e., the pixel-wise product of the attention map and intermediate features.

\section{Generative Model}
Going from natural language phrases to color and being able to synthesize color profiles of unseen compositions is a challenging task. There have been prior efforts to learn this mapping from text modality alone~\cite{maheshwari2020learning,monroe2017colors}, but the perception of color is inherently rooted in the visual domain. To tackle this problem, we adopt a multimodal approach to learn color representations of (attribute, object) pairs, ensuring that images are required only for training, not for inference. We propose a generative model that learns compositionality and context in color space. The generator predicts plausible color profiles conditioned on the text embedding, while the discriminator attempts to distinguish between real color profiles (from images) and generator outputs. Our approach is motivated by the recent success of adversarial examples in zero-shot compositional learning for image classification~\cite{wei2019adversarial,wang2019task}.

\begin{figure}[!h]
\centering
\includegraphics[width=\columnwidth]{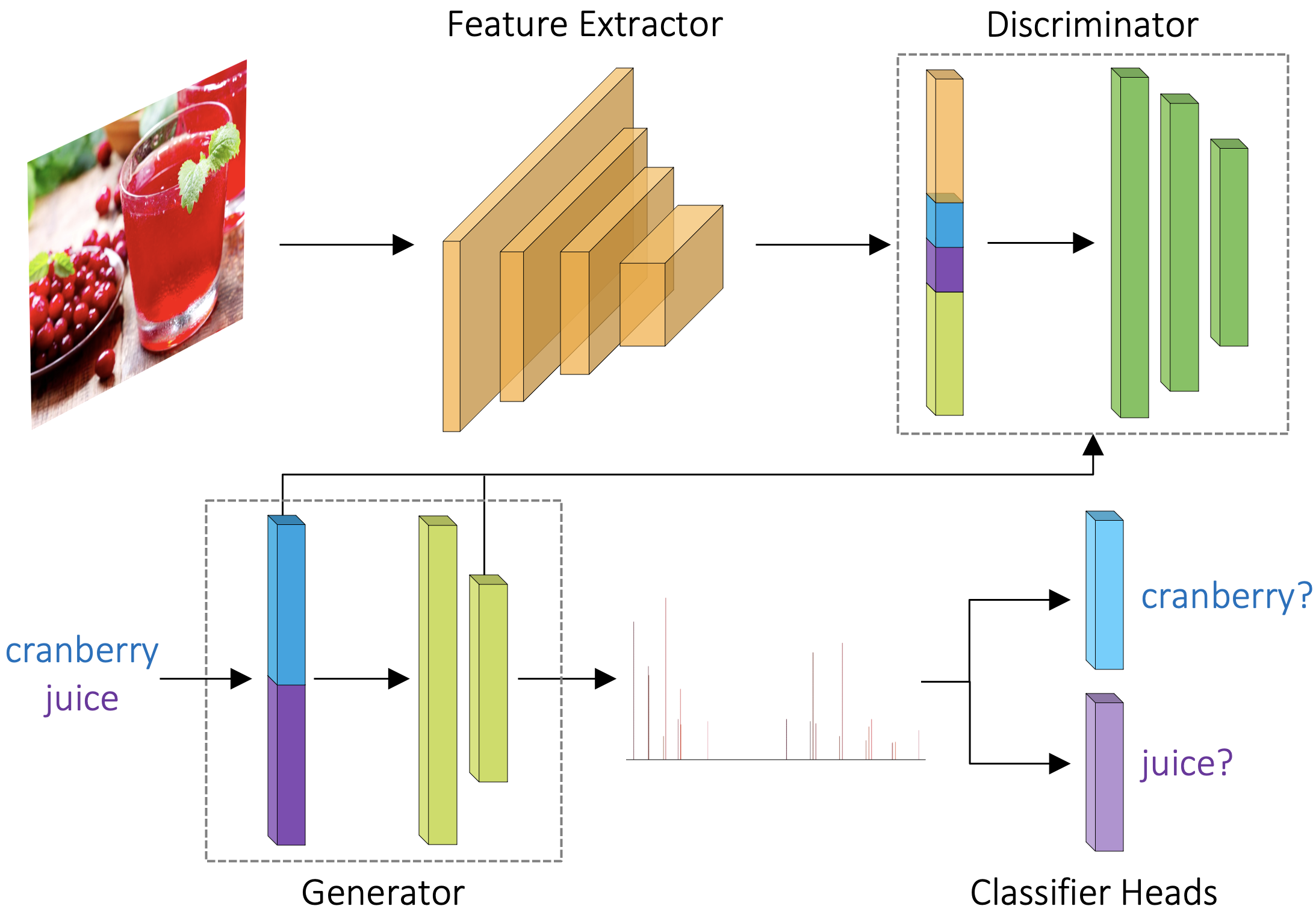}
\caption{Generative adversarial network for learning color representations of (attribute, object) text phrases.}
\label{model}
\end{figure}

The generator network uses word embeddings to capture the initial context of the text, followed by fully connected layers with ReLU activation, and finally softmax to return the color embedding. It is key to note that we use different trainable embedding matrices for attributes and objects because the same word can have multiple interpretations based on its linguistic type, e.g., \textit{sea} in \textit{`deep sea'} is an object but plays the role of an attribute in \textit{`sea blue'}). The image modality is only input to the discriminator network, and it provides feedback to the generator via the adversarial loss. The discriminator is another neural network that takes as input the text embeddings, pretrained image features, and a color profile; and predicts a real versus fake score between $[0,1]$. The overall network architecture to map text to color profiles is shown in Figure~\ref{model}, and individual components are detailed next.

\subsection{Training Objective}
We train our GAN model using a modified Least Squares GAN objective~\cite{mao2017least}. Mathematically, the generator $G$ and discriminator $D$ objectives are given by
\begin{align*}
\mathcal{L}_G = \mathbb{E}_{\mathbf{a,o}}\left[\Big(D\big(G(\mathbf{a,o}) \hspace{0.5mm}|\hspace{0.5mm} \mathbf{a, o}\big) - 1\Big)^{2}\right] + \uplambda_\text{color}\mathcal{L}_\text{color} + \uplambda_\text{cls}\mathcal{L}_\text{cls}
\end{align*}
\begin{align*}
\mathcal{L}_D = \hspace{2mm} & \mathbb{E}_{\mathbf{x}_{a,o}} \left[\Big(D\big(\mathbf{x}_{a,o} \hspace{0.5mm}|\hspace{0.5mm} \mathbf{a, o}\big) - 1\Big)^{2}\right] \\
& \hspace{5mm} + \mathbb{E}_{\mathbf{a,o}}\left[\Big(D\big(G(\mathbf{a,o}) \hspace{0.5mm}|\hspace{0.5mm} \mathbf{a, o}\big) \Big)^{2}\right] + \uplambda_\text{mis}\mathcal{L}_\text{mis}
\end{align*}
where $D\big(\mathbf{x}_{a,o} \hspace{0.5mm}|\hspace{0.5mm} \mathbf{a, o}\big)$ represents the output score of the discriminator on seeing a real color profile, and $D\big(G(\mathbf{a,o}) \hspace{0.5mm}|\hspace{0.5mm} \mathbf{a, o}\big)$ is for the generated color profile. Thus, the discriminator tries to minimize the score for a real color profile $\mathbf{x}_{a,o}$ which is sampled from the set of images corresponding to the composition $(\mathbf{a,o})$, and maximize the score of the output $G(\mathbf{a,o})$ by the generator. The generator, on the other hand, is trained to maximize the discriminator's score for its generated output.

\textbf{Color Loss $\mathcal{L}_\text{color}$} : This measures the distance between the predicted color representations and true ones from images, and is used to guide the training of the generator. We experiment with different contrastive losses which are described next. $\mathbf{x_{a,o}}$ denotes the attention-weighted color profile sampled uniformly at random from the set of all images for class $(\mathbf{a,o})$, and $\mathbf{\hat{x}}_{a,o}$ is the model prediction $G(\mathbf{a,o})$.

\begin{enumerate}
\itemsep0.5em
    \item \textbf{L2 Loss}: This is a simple euclidean distance between the color profiles.
    \begin{equation*}
    \mathcal{L}_{\ell2} \left(\mathbf{\hat{x}}_{a,o}, \mathbf{x}_{a,o}\right) = \lVert\hspace{0.75mm} \mathbf{\hat{x}}_{a,o} - \mathbf{x}_{a,o} \hspace{0.75mm}\rVert_2
    \end{equation*}

    \item \textbf{Triplet Loss}: Inspired from the widely-used contrastive paradigms in the vision community, triplet or margin loss~\cite{weinberger2006distance,schroff2015facenet} takes a positive sample $\mathbf{x}_{a,o}$ of the same class and a negative one $\mathbf{x}_{\bar{a},\bar{o}}$, and tries to bring the anchor $\mathbf{\hat{x}}_{a,o}$ close to the positive and far from the negative.
    \begin{equation*}
    \begin{aligned}
    \hspace{8mm} \mathcal{L}_\text{triplet} & \left(\mathbf{\hat{x}}_{a,o}, \mathbf{x}_{a,o}, \mathbf{x}_{\bar{a},\bar{o}}\right) = \\
    & \max\Big(0, \mathcal{L}_{\ell2}\left(\mathbf{\hat{x}}_{a,o}, \mathbf{x}_{a,o}\right) - \mathcal{L}_{\ell2}\left(\mathbf{\hat{x}}_{a,o}, \mathbf{x}_{\bar{a},\bar{o}}\right) + m \Big)
    \end{aligned}
    \end{equation*}
    where the negative histogram  $\mathbf{x}_{\bar{a},\bar{o}}$ is randomly sampled from any other class $(\mathbf{\bar{a},\bar{o}})$, and $m$ is the margin hyperparameter.

    \item \textbf{Quintuplet Loss}: This extends~\cite{wei2019adversarial} the triplet loss by considering multiple task specific negatives. It considers one negative $\mathbf{x}_{\bar{a},\bar{o}}$ belonging to the $(\mathbf{\bar{a},\bar{o}})$ class and two semi negatives $\mathbf{x}_{a,\bar{o}}$ and $\mathbf{x}_{\bar{a},o}$, which have either the same attribute $(\mathbf{a,\bar{o}})$ or the same object $(\mathbf{\bar{a},o})$ as the anchor. The loss is a weighted sum of $3$ triplet components, given by
    \begin{equation*}
    \begin{aligned}
    \mathcal{L}_\text{quintuplet} \left(\mathbf{\hat{x}}_{a,o}, \mathbf{x}_{a,o}, \mathbf{x}_{\bar{a},\bar{o}}, \mathbf{x}_{a,\bar{o}}, \mathbf{x}_{\bar{a},o} \right) = \hspace{5mm} \\
    \uplambda_1 \mathcal{L}_\text{triplet} \left(\mathbf{\hat{x}}_{a,o}, \mathbf{x}_{a,o}, \mathbf{x}_{\bar{a},\bar{o}}\right) \\
    + \hspace{1.5mm} \uplambda_2 \mathcal{L}_\text{triplet} \left(\mathbf{\hat{x}}_{a,o}, \mathbf{x}_{a,o}, \mathbf{x}_{a,\bar{o}}\right) \\
    + \hspace{1.5mm} \uplambda_3 \mathcal{L}_\text{triplet} \left(\mathbf{\hat{x}}_{a,o}, \mathbf{x}_{a,o}, \mathbf{x}_{\bar{a},o}\right)
    \end{aligned}
    \end{equation*}
    where weight hyperparameters are such that $\uplambda_1 > \uplambda_2 = \uplambda_3$.
\end{enumerate}
This additional color loss in the generator’s objective helps in combating mode collapse and stabilizing training.

\textbf{Classification Loss $\mathcal{L}_\text{cls}$} : The generator output is passed through two different classifier heads - one for attribute, and other object. Simultaneously, a cross-entropy loss component is added to the generator objective.
\begin{align*}
\mathcal{L}_\text{cls} = \hspace{0.5mm}-\hspace{0.5mm} \mathbb{E}_{\mathbf{a,o}}\left[ \log P_\mathbf{a}\hspace{0.5mm} \Big(\mathbf{a} \hspace{0.5mm}|\hspace{0.5mm} \mathbf{\hat{x}}_{a,o} \Big) \right] \hspace{0.5mm}-\hspace{0.5mm} \mathbb{E}_{\mathbf{a,o}}\left[ \log P_\mathbf{o}\hspace{0.5mm} \Big(\mathbf{o} \hspace{0.5mm}|\hspace{0.5mm} \mathbf{\hat{x}}_{a,o} \Big) \right]
\end{align*}
where $P_\mathbf{a}$ and $P_\mathbf{o}$ denote the conditional probability of the respective classifiers making the right prediction. This is done to incorporate feedback from the closely-related task of color naming~\cite{monroe2016learning,monroe2017colors,mcmahan2015bayesian}. It has also been shown to improve the generator's ability to generalize over unseen compositions of (attribute, object) pairs.

\textbf{Mismatch Loss $\mathcal{L}_\text{mis}$} : This term extends the conditional GAN loss~\cite{mirza2014conditional} by encouraging the discriminator to classify mismatched combinations of generated color profiles and text inputs as fake~\cite{reed2016generative}. Here, the discriminator minimizes the score given to the combination of the real color profile $\mathbf{x}_{a,o}$ and the mismatched composition $(\mathbf{\bar{a}, \bar{o}})$, forcing the discriminator to explicitly identify class mismatch in addition to the traditional real/fake distinction.
\begin{align*}
\mathcal{L}_\text{mis} = \mathbb{E}_{\mathbf{a,o}}\left[\Big(D\big(\mathbf{x}_{a,o} \hspace{0.5mm}|\hspace{0.5mm} \mathbf{\bar{a}, \bar{o}}\big)\Big)^{2}\right]
\end{align*}
Since our setup is a conditional GAN, the discriminator needs to evaluate the conditioning constraint of the generator's output on the input text. In turn, this feedback to the generator also ensures that the predicted color profiles are not only plausible but also correlated with the text.

Note that all loss components use the attention mechanism described before for color profiles obtained from images. We follow an alternate training strategy, wherein the generator is trained for $K$ epochs, followed by discriminator training for $K$ epochs, and so on. This switching is done to stabilize learning - giving both the networks sufficient iterations to train smoothly before the adversarial component drives training and further improves performance.

\begin{table*}[!h]
\centering
\caption{Evaluation of different text-to-color models on both seen and unseen compositions of (attribute, object) pairs.}
\label{trainingResults}
\begin{tabular}{lr c cc c cc c}
\toprule
\multicolumn{2}{c}{\textbf{Model}} && \multicolumn{2}{c}{\textbf{Seen Comp.} {\scriptsize ($\times 10^{-3}$)}} && \multicolumn{2}{c}{\textbf{Unseen Comp.} {\scriptsize ($\times 10^{-3}$)}} & \multirow{2}{*}{\textbf{FID}} \\
\cmidrule{1-2}\cmidrule{4-5}\cmidrule{7-8}
\multicolumn{1}{c}{\textbf{Network}} & \multicolumn{1}{c}{\textbf{$\mathcal{L}_\text{color}$}} && \textbf{Macro L2} & \textbf{Micro L2} && \textbf{Macro L2} & \textbf{Micro L2} & \\
\midrule
\multirow{3}{*}{Label Embed} & $+ \mathcal{L}_{\ell2}$ && 0.244 & 0.865 && 0.233 & 0.847 & 0.615 \\
& $+ \mathcal{L}_\text{triplet}$ && 0.581 & 1.089 && 0.578 & 1.076 & 0.368 \\
& $+ \mathcal{L}_\text{quintuplet}$ && 0.624 & 1.111 && 0.623 & 1.098 & 0.366 \\[0.8ex]
\hdashline\noalign{\vskip 1ex}
\multirow{3}{*}{Ours} & $+ \mathcal{L}_{\ell2}$ && 0.155 & 0.775 && 0.355 & 0.936 & 0.391 \\
& $+ \mathcal{L}_\text{triplet}$ && 0.460 & 0.976 && 0.716 & 1.189 & 0.208 \\
& $+ \mathcal{L}_\text{quintuplet}$ && 0.575 & 1.059 && 0.800 & 1.262 & 0.233 \\[0.8ex]
\hdashline\noalign{\vskip 1ex}
\multirow{3}{*}{Ours $- \mathcal{L}_\text{mis}$} & $+ \mathcal{L}_{\ell2}$ && 0.162 & 0.791 && 0.360 & 0.949 & 0.383 \\
& $+ \mathcal{L}_\text{triplet}$ && 0.481 & 0.997 && 0.730 & 1.211 & 0.251 \\
& $+ \mathcal{L}_\text{quintuplet}$ && 0.578 & 1.061 && 0.759 & 1.228 & 0.225 \\[0.8ex]
\hdashline\noalign{\vskip 1ex}
\multirow{3}{*}{Ours $- \mathcal{L}_\text{cls}$} & $+ \mathcal{L}_{\ell2}$ && 0.130 & 0.759 && 0.340 & 0.917 & 0.428 \\
& $+ \mathcal{L}_\text{triplet}$ && 0.451 & 0.955 && 0.707 & 1.172 & 0.203 \\
& $+ \mathcal{L}_\text{quintuplet}$ && 0.560 & 1.009 && 0.805 & 1.230 & 0.205 \\[0.8ex]
\hdashline\noalign{\vskip 1ex}
\multirow{3}{*}{Ours $- \mathcal{L}_\text{mis} - \mathcal{L}_\text{cls}$ \hspace{2mm}} & $+ \mathcal{L}_{\ell2}$ && 0.137 & 0.763 && 0.335 & 0.915 & 0.445 \\
& $+ \mathcal{L}_\text{triplet}$ && 0.466 & 0.970 && 0.721 & 1.187 & 0.223 \\
& $+ \mathcal{L}_\text{quintuplet}$ && 0.557 & 1.023 && 0.801 & 1.238 & 0.229 \\[0.8ex]
\bottomrule
\end{tabular}
\end{table*}

\subsection{Results and Evaluation}
We now describe our experimental setup, the baseline models, and present qualitative and quantitative evaluation of our approach.

\textbf{Implementation Details}: We set the hyperparameters in the dataset curation pipeline as $f = 10$, $t_a = 5$ and $t_o = 5$, and obtain the final set of (attribute, object) pairs by running the bipartite graph filtering for $h=2$ hops. We split the set of all (attribute, object) pairs in the ratio $70:15:15$ for training, validation and testing respectively. This entails that all images of a given class fall into the same set. Therefore, the compositions of the test set are never seen by the model and corresponds to zero-shot learning.

For the generator, we first embed the (attribute, object) pair into a trainable $300$ dimension embedding using GloVe vectors~\cite{glove}. This is followed by separate FC layers of size $400$ each. The attribute and object embeddings are then concatenated and passed through another fully connected network with $1000$ and $1504$ hidden units respectively. We add dropout~\cite{dropout} to the first hidden layer, with the dropout rate set to $0.4$. The embedding is finally normalized using softmax (at the resolution level, i.e., for the first $504$ and last $1000$ bins separately), resulting in the color representation. Both the attribute and object classifiers are a single linear layer with softmax activation to predict a probability distribution over the set of all attribute and object classes. The discriminator concatenates the $3$ inputs - $800$ length text conditioning, $1504$ length colour representation, $2048$ length image feature computed using a pretrained ResNet model~\cite{resnet}. It is passed through $2$ hidden layers with ReLU activation of $4000$ and $2000$ units respectively. Lastly, a linear layer predicts a real/fake score for the conditioned input.

The end-to-end network is trained using RMSprop optimizer for $300$ epochs and a batch size of $64$. We use a learning rate of $10^{-4}$ for the generator and classifier, and $10^{-5}$ for the discriminator. The hyperparameters are set as follows for all involved experiments: $\uplambda_\text{cls}=0.05$, $\uplambda_\text{mis}=1$ in the overall objective; $\uplambda_\text{color}=0.6$ for L2 loss; $\uplambda_\text{color}=2$, $m=1$ for triplet loss; $\uplambda_\text{color}=0.1$, $\uplambda_1=1$, $\uplambda_2=0.3$, $\uplambda_3=0.3$ for quintuplet loss; and $K=10$ for alternate training.

\textbf{Baseline}: We evaluate our model against a discriminative baseline that learns to compose color profiles from constituent word embeddings. The architecture of this model is the same as that of the generator, allowing for a fair comparison of the computed color representations. We refer to this baseline as \textit{Label Embed}~\cite{nagarajan2018attributes}. This network is trained using the ground-truth color profiles from images of the corresponding class, and $\mathcal{L}_\text{color}$ as the objective.

\begin{figure}[t]
\centering
\includegraphics[width=\linewidth]{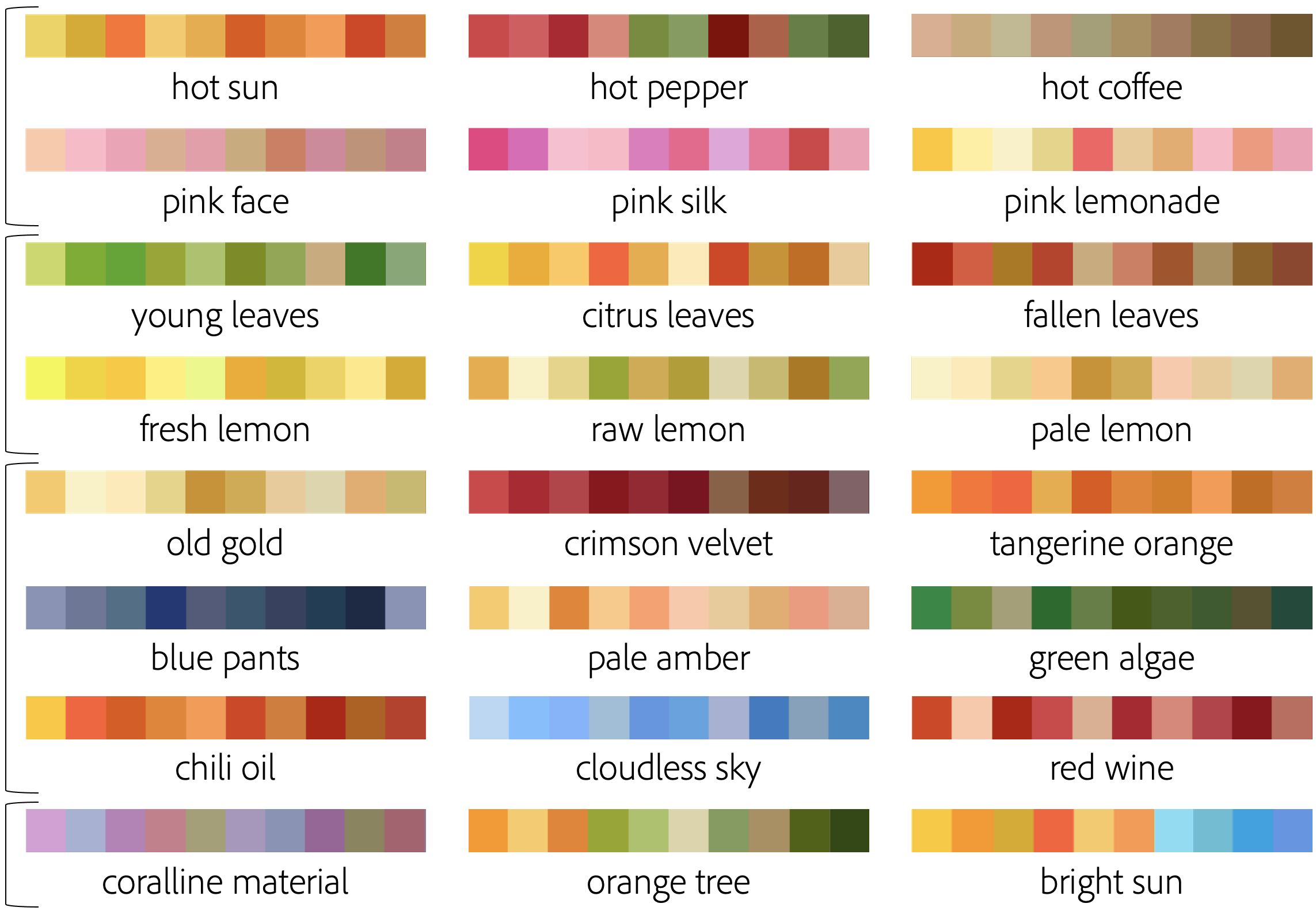}
\caption{Generated color palettes. Rows 1,2 depict how an attribute alters different objects, while rows 3,4 show the same object with varying attribute types. Row 5-7 comprises of unseen combinations of (attribute, object) pairs. Row 8 illustrates complex color profiles with multiple color intents.}
\label{colorProfiles}
\end{figure}

\textbf{Evaluation}: We evaluate the predicted color representations using the following metrics (i) \textit{Macro L2} which measures the L2 distance between model predictions and average histogram across all instances of all (attribute, object) classes, (ii) \textit{Micro L2} which measures the average L2 distance between model predictions and individual image histograms of the corresponding class and then averaged across classes, and (iii) \textit{Fr\'echet Inception Distance (FID)}~\cite{fid} to estimate the quality of our conditional GAN network. The FID is a comparison between statistics of the two distributions - color profiles generated by model versus real ones from images. We report these metrics at two levels - for seen (attribute, object) compositions of the training set and unseen compositions (zero-shot) of the test set. We also perform ablation experiments to study the effectiveness of different loss components while training the GAN.

The numbers are shown in Table~\ref{trainingResults} from which we can make the following observations. All variants of the generative model consistently outperform Label Embed for seen compositions. There is a drop in performance for unseen compositions, which is natural. Comparing across different $\mathcal{L}_\text{color}$ objectives, L2 loss leads to the lowest Macro and Micro L2 losses, primarily due to the parallels in the training and evaluation metrics. But the importance of contrastive alternatives (Triplet and Quintuplet losses) is evident from the FID scores. The FID metric highlights the main advantage of our generative setup - the predicted color profiles are statistically much closer to real color profiles obtained from images. This is a direct consequence of our discriminator network which utilises the visual modality for enhanced training. Classification loss $\mathcal{L}_\text{cls}$ leads to a slight increase in the L2 metrics, but it enables the learning of realistic color profiles, noticeable from the lower FID scores.

We further provide qualitative evidence in Figure~\ref{colorProfiles}. Our model captures context - effect of an attribute on different objects - for abstract concepts like `hot', and explicit color indicators like `pink'. Our model also learns the notion of composition - how different attributes modify the same object. The color of `young leaves' is rich in green whereas `fallen leaves' are represented well in the brown-to-red spectrum and `citrus leaves' are more yellowish. A similar argument follows for the object `lemon' and modifiers such as `fresh', `raw' and `pale'. It also learns meaningful colors for unseen combinations of (attribute, object) pairs. Rows 5 through 7 demonstrate effective zero-shot learning as the generated color profiles reasonably capture the semantics of the text. Another interesting behavior is its ability to highlight multiple color shades. For `bright sun', it has learned to depict a golden yellow sun in a blue sky. Similarly, the model predicts multiple dominant color for the phrases `coralline material' and `orange tree'. All these example texts were obtained via the proposed dataset curation logic.

\section{Color for Downstream Tasks}
In the previous section, we described our proposed generative model to go from (attribute, object) text phrases to color representations and palettes. We believe that there are multiple downstream applications where such mapping can prove useful. Some tasks include cross-modal ranking~\cite{maheshwari2020learning,cox2000bayesian}, language-driven image editing and manipulation~\cite{chen2018language,heer2012color}, as well as image colorization~\cite{bahng2018coloring,manjunatha2018learning,zou2019language}. Our model can be used to capture users' color intent in a much more intuitive manner using natural language. For example, the phrase `dry leaves' conveys rich semantics, while simultaneously eliminating the cumbersome process of selecting RGB values manually. While we acknowledge that we have focused on a very specific linguistic structure for the phrases, i.e adjective and noun combinations, they form a common structure of generic text inputs.

\begin{figure*}[!h]
\centering
\includegraphics[width=0.95\linewidth]{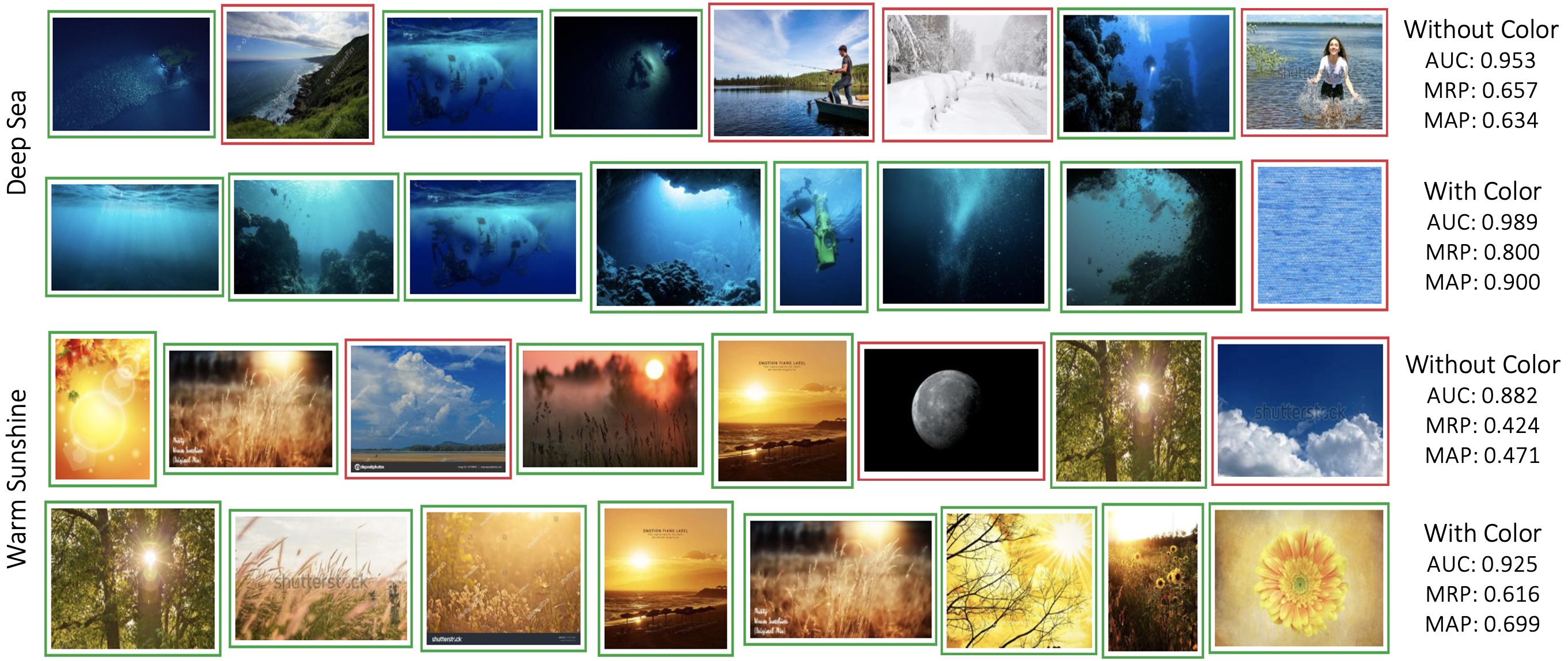}
\caption{Qualitative results depicting the benefits of color-centric features in cross-modal retrieval. Ranking results for two exemplar queries along with retrieval metrics. Images bound in green belong to the query class and images in red are irrelevant.}
\label{rankingComparison}
\end{figure*}

In this work, we focus on the task of cross-modal retrieval, specifically the role that color plays in it. We confine to textual queries with the (attribute, object) bigram structure. We design a relevance matching model that takes such a textual phrase and an image as input, and produces a relevance score between them. We train a standard multi-modal network~\cite{frome2013devise} for shared representation learning. This network is trained using a contrastive approach where the model learns to rank relevant images higher than irrelevant ones. The overall objective is to maximise the difference of scores between the positive (relevant) and negative (not relevant). Mathematically,
\begin{equation*}
    \mathcal{L}_\text{ranker} = -\, \mathbb{E}_{\mathbf{a,o}}\,\left[ \sigma \Big( R\big(\,\mathbf{a, o} \,, \mathcal{\mathbf{I}}_{a,o}\,\big) - R\big(\,\mathbf{a, o} \,, \mathcal{\mathbf{I}}_{\bar{a},\bar{o}}\,\big) \Big) \right]
\end{equation*}
where $\sigma(x)$ is the sigmoid activation function and $R$ is the ranker. $R\big(\,\mathbf{a, o} \,, \mathcal{\mathbf{I}}_{a,o}\,\big)$ denotes the score predicted by the ranker for text $(\mathbf{a,o})$ and positive image $\mathcal{\mathbf{I}}_{a,o}$ from the same class, and $ R\big(\,\mathbf{a, o} \,, \mathcal{\mathbf{I}}_{\bar{a},\bar{o}}\,\big)$ is the ranker output for text $(\mathbf{a,o})$ and negative image $\mathcal{\mathbf{I}}_{\bar{a},\bar{o}}$ chosen randomly from any other class $(\mathbf{\bar{a},\bar{o}})$. This is inspired from the RankNet loss~\cite{burges2005learning} which uses clicked-versus-not labels to improve search ranking from user behavioral data. Finally, the retrieval performance of this model is evaluated in terms of ranking the visual assets against the text queries. We also utilize and evaluate the model as a classifier, by scoring a given image against a fixed enumeration of (attribute, object) textual phrases and measuring its ability in scoring the correct one the highest.

\begin{table*}[!h]
\centering
\caption{Evaluation of image-text relevance matching models on cross-modal retrieval and image classification tasks. We consider different model variations in addition to base features, and they are as follows: (1a) no color-specific features (1b) only image color (1c) image color and ground-truth text color (2) image color and generated text color.}
\label{ranker}
\begin{tabular}{lr c ccc c ccc}
\toprule
\multirow{3}{*}{} &&& \multicolumn{3}{c}{\textbf{Retrieval at $k=5$}} && \multicolumn{3}{c}{\textbf{Top-$20$ Classification Accuracy}} \\
\cmidrule{4-6}\cmidrule{8-10}
&&& \textbf{AUC} & \textbf{MRP} & \textbf{MAP} &&  \textbf{Pair} & \textbf{Attribute} & \textbf{Object} \\
\midrule
\multicolumn{10}{c}{\textit{(1) Using Ground Truth Color Representations}} \\
\midrule
\multicolumn{2}{l}{Base Features} && 0.837 & 0.510 & 0.534 && 7.521 & 11.403 & 11.200 \\
\multicolumn{2}{l}{Base Features + Image Color} && 0.866 & 0.570 & 0.597 && 13.521 & 16.983 & 15.720 \\
\multicolumn{2}{l}{Base Features + Image Color + Text Color} && 0.905 & 0.656 & 0.692 && 22.644 & 27.477 & 23.812 \\
\midrule
\multicolumn{10}{c}{\textit{(2) Using Generated Color Representations for Text Color}} \\
\midrule
\multirow{3}{*}{Ours} & $+ \mathcal{L}_{\ell2}$ && 0.858 & 0.573 & 0.605 && 16.236 & 23.839 & 26.758 \\
& $+ \mathcal{L}_\text{triplet}$ && 0.862 & 0.577 & 0.607 && 19.576 & 22.753 & 25.210 \\
& $+ \mathcal{L}_\text{quintuplet}$ && 0.856 & 0.567 & 0.597 && 15.367 & 19.074 & 16.969 \\[0.8ex]
\hdashline\noalign{\vskip 1ex}
\multirow{3}{*}{Ours $- \mathcal{L}_\text{mis}$} & $+ \mathcal{L}_{\ell2}$ && 0.861 & 0.577 & 0.607 && 15.408 & 22.631 & 28.780 \\
& $+ \mathcal{L}_\text{triplet}$ && 0.862 & 0.576 & 0.604 && 12.367 & 15.381 & 16.915 \\
& $+ \mathcal{L}_\text{quintuplet}$ && 0.860 & 0.569 & 0.598 && 12.449 & 17.635 & 15.734 \\[0.8ex]
\hdashline\noalign{\vskip 1ex}
\multirow{3}{*}{Ours $- \mathcal{L}_\text{cls}$} & $+ \mathcal{L}_{\ell2}$ && 0.859 & 0.573 & 0.606 && 22.821 & 31.387 & 30.627 \\
& $+ \mathcal{L}_\text{triplet}$ && 0.854 & 0.568 & 0.594 && 11.987 & 18.666 & 26.622 \\
& $+ \mathcal{L}_\text{quintuplet}$ && 0.856 & 0.561 & 0.590 && 20.621 & 25.061 & 24.368 \\[0.8ex]
\hdashline\noalign{\vskip 1ex}
\multirow{3}{*}{Ours $- \mathcal{L}_\text{mis} - \mathcal{L}_\text{cls}$ \hspace{2mm}} & $+ \mathcal{L}_{\ell2}$ && 0.856 & 0.574 & 0.602 && 17.553 & 24.083 & 28.740 \\
& $+ \mathcal{L}_\text{triplet}$ && 0.863 & 0.577 & 0.604 && 19.902 & 24.830 & 26.364 \\
& $+ \mathcal{L}_\text{quintuplet}$ && 0.860 & 0.570 & 0.598 && 14.417 & 21.327 & 29.459 \\[0.8ex]
\bottomrule
\end{tabular}
\end{table*}

We experiment with different combinations of input features, specifically with and without explicit color information, and attribute the gain in performance to the use of color representations. The baseline model uses only pretrained word and image embeddings (collectively termed as \textit{Base Features}) for the text and image networks respectively. We next incrementally provide the model with color representations of images (termed as \textit{Image Color}). Lastly, we build a model that also uses the output of our text-to-color model, i.e, color representations of text phrases (termed as \textit{Text Color}). Here, we first wish to study the effect of text color in ranking in isolation, i.e., independent of the performance of our GAN setup. So we define a ground-truth color representation for (attribute, object) bigrams as the average over all image histograms belonging to that class. This acts as an upper limit on the ranker performance when text color is added. Then, we use the color profiles generated from our model and evaluate the ranker and GAN together. Since our goal is to evaluate a color-centric feature, we work in a controlled setup – a simple baseline that achieves reasonable accuracy. We do not use additional metadata such as image tags or captions, which are otherwise common in image retrieval systems.

\textbf{Implementation Details}: As before, base features are extracted from pretrained ResNet~\cite{resnet} and text features are a concatenation of individual GloVe embeddings~\cite{glove}. All ground-truth color features are obtained from attention-weighted LAB space color histograms. The ranker comprises of $2$ hidden layers with ReLU activation of $1024$ and $512$ units respectively, after which a linear layer returns a scalar value for the image-text relevance score.

\textbf{Evaluation}: Our retrieval setup is as follows - For a given $\mathbf{a, o}$ pair, we consider the set of all relevant images $| \mathcal{\mathbf{I}}_{a,o} | = n_{a,o}$ of that class and randomly sample $k * n_{a,o}$ irrelevant images from the dataset, where $k$ is a hyperparameter. We consider a range of values of $k$ and measure the retrieval performance of our model using standard IR metrics – Area Under the ROC curve (AUC), Mean R-Precision (MRP) and Mean Average Precision (MAP). As $k$ increases, the task difficulty also increases as the model now has to differentiate between relevant and irrelevant images for a text query from a much larger pool.

For image classification, we consider all the pairs in the dataset and assign relevance scores to each pair using the relevance matching model. We then calculate the top-$N$ classification accuracy as the percentage of images for which the correct class appeared in the top $N$ predictions made by the model. We also extend this model to attribute only and object only classification tasks. For this, we define the attribute only relevance as the average across all pairs with that attribute, and similarly for objects.

Table~\ref{ranker} summarizes the retrieval results at $k=5$ and Top-$20$ classification accuracies. In the set of experiments in Part (1), the \textit{Text Color} is defined as the average color profile of all images relevant for that text query. It is evident from the metrics that incorporating color of both modalities outperforms the other model variations - no color features or only color in images. It is worth noting that there is a consistent improvement in performance by adding \textit{Text Color} to \textit{Base Features + Image Color}, and the model is able to achieve an AUC of $\sim0.9$. This indicates the significance of having a color representation for not just images, but text modality too, and further corroborates our motivation.

In Part (2), we evaluate our GAN architecture in a ranking and classification context by using the model predictions as \textit{Text Color} features for relevance matching. While the metrics are marginally lower compared to the use of ground-truth for \textit{Text Color}, this is a natural outcome of the use of model predictions. Even then, all variations of the GAN objective lead to better results than the baseline model which uses just \textit{Base Features}. These results ascertain the viability of our generative model for text-to-color prediction, and the promising use of color in various downstream applications.

In figure~\ref{rankingComparison}, we provide a side-by-side comparison of image retrieval using a model that does not use explicit color information versus one that does. Consider the query `deep sea' - the ranker with color has captured the intuition that ``deep'' turns the shade of water darker. Similarly, for the query `warm sunshine', the ranker with color can retrieve more yellowish images, while the baseline ranker (which does not use color features) fetches several pictures of a blue sky. The same can also be observed from the retrieval metrics where adding color leads to improved performance.

\section{Conclusion and Discussion}
In this paper, we have considered the task of generating color representations from text. We focused our attention on textual phrases with an (attribute, object) structure, motivated by their common occurrence in user queries of an image search engine. In addition, as seen using examples throughout the paper, such phrases exhibit much visual diversity, especially on the color axis.

We have described a dataset curation strategy that we believe is a useful general workflow for studying the mapping of text to other visual axes, like texture and aesthetics. We propose a GAN architecture that when trained over the compiled dataset generates intuitive output. We have also conducted a quantitative evaluation via the use of multiple types of metrics - general notions of difference between embeddings, color-specific distance functions, as well as the Fr\'echet Inception Distance (FID) specifically for use with GAN outputs. While the non-generative baseline performs equivalently on other metrics, in terms of the predicted color representations, the GAN has a better FID score indicating that the colors it produces are more realistic in general.

While the task of producing color from text has been explored in prior work, our specific focus was on the interplay between context and compositionality. When an adjective is used alongside a noun, it modifies the visual representation of the corresponding object. Encouraging compositionality (separating the network connections in the early layers of the model) aids in being able to combine previously seen primitives (i.e., attribute and object words) into novel concepts. Parts of our evaluation were directed at this zero-shot setting. In addition, we have illustrated the important role of color in the downstream task of image retrieval, and the improvements via the use of the color predictions from our model.

In future work, we wish to expand to a wider class of linguistic structures observed in phrases. It is common to have multiple attributes for a given object (e.g. 'round metallic bottle'). Modeling the compounded effect of the attributes on the object, in a manner not limited by their number, would be an interesting task. In addition, specifically focusing on the visual axis of color, some modifiers tend to be commonly observed such as `dark' and `deep'. Similar to the earlier point, attributes can be cascaded (`very dark blue') offering a very rich vocabulary for the text phrases. Following the pipeline described in this paper - which includes a way to enumerate a list of example phrases of that structure, obtaining ground truth derived from images, designing and training models that go from text to color - offers scope for multiple investigations.

\bibliographystyle{ACM-Reference-Format}
\bibliography{main}

\end{document}